\documentclass[10pt,twocolumn,letterpaper]{article}
\usepackage{booktabs}
\usepackage{multirow}
\usepackage{graphicx}

\usepackage{cvpr}      % To produce the REVIEW version
\usepackage{algorithm}
\usepackage[noend]{algpseudocode}
\algrenewcommand\algorithmicrequire{\textbf{Input:}}
\algrenewcommand\algorithmicensure{\textbf{Output:}}
\definecolor{cvprblue}{rgb}{0.21,0.49,0.74}
\usepackage[pagebackref,breaklinks,colorlinks,allcolors=cvprblue]{hyperref}

\def\confName{CVPR}
\def\confYear{2027}

\title{RotGuard: Identity Protection in Personalized Diffusion Models via Rotation-Guided Geometry Disruption}

\author{
    Lizhi Xiong$^{1}$,
    Yang Li$^{1}$,
    Jun Li$^{2}$,
    Ziqiang Li$^{1}$,
    Zhangjie Fu$^{1}$\\
    $^{1}$Nanjing University of Information Science and Technology, Nanjing, China\\
    $^{2}$Hefei University of Technology, Hefei, China\\
    {\tt\small lzxiong16@163.com, ly3071319579@gmail.com, lijuun@yeah.net,}\\
    {\tt\small iceli@mail.ustc.edu.cn, fzj@nuist.edu.cn}
    }
\begin{document}
\maketitle

\begin{abstract}
%Personalized diffusion models enable identity-preserving image generation but raise privacy risks by facilitating unauthorized identity replication. Existing proactive defenses perturb latent representations, image features, or denoising trajectories, yet they fail to sufficiently disrupt the geometric structure of identity features. We investigate this issue through cross-attention Query (Q) features and show that disrupting their relative angular relationships weakens identity preservation more effectively than changing feature magnitudes or applying shared orthogonal rotations. Further analysis reveals that existing defenses largely preserve these angular relationships, allowing identity-related information to remain exploitable during generation. Based on this finding, we propose \textbf{RotGuard}, a geometry-guided framework integrating Identity Feature Space Repulsion (IFSR), Query Structure Disruption (QSD), and Conditional Denoising Shift (CDS) to interfere with identity feature extraction, fusion, and identity-guided denoising trajectories. By jointly optimizing these components on a surrogate model, RotGuard generates protected images transferable across personalization architectures. Extensive experiments on U-Net- and DiT-based models demonstrate its effectiveness, cross-architecture transferability, and robustness against post-processing and purification attacks.
Personalized diffusion models enable identity-preserving image generation but raise privacy concerns due to unauthorized identity replication. Existing proactive defenses perturb feature representations or denoising processes, yet their effectiveness and transferability remain limited. We investigate the role of feature geometry in identity preservation and find that disrupting the relative angular relationships among cross-attention Query (Q) features substantially weakens identity preservation, whereas modifying feature magnitudes or global orientations has limited effects. Further analysis shows that existing defenses largely preserve these geometric relationships. Motivated by these findings, we propose \textbf{RotGuard}, a geometry-guided framework for transferable identity protection. RotGuard integrates two complementary rotation-based components: \textbf{Query Structure Disruption (QSD)}, which disrupts Q-feature geometry to interfere with identity feature fusion, and \textbf{Conditional Denoising Shift (CDS)}, which alters identity-guidance geometry to redirect denoising trajectories. An auxiliary \textbf{Identity Feature Space Repulsion (IFSR)} component further weakens identity feature extraction. Jointly optimized on a surrogate model, RotGuard generates protected reference images transferable across personalization architectures. Extensive experiments on U-Net- and DiT-based models demonstrate its effectiveness, cross-architecture transferability, and robustness against post-processing and purification attacks.
\end{abstract}    
\section{Introduction}\label{sec:intro}
%In recent years, the rapid development of text-to-image diffusion models\cite{LDM,2} has made personalized image generation an increasingly active research area. Personalized diffusion models can extract subject-specific identity features from a small number of reference images, enabling high-fidelity personalized generation\cite{blip,face0,rb}. Existing personalization methods can be broadly divided into \textbf{training-based} and \textbf{training-free} approaches. Early training-based methods\cite{textualinversion,dreambooth,lora,custom} learn subject-specific identity information by fine-tuning pretrained diffusion models on a set of reference images. However, as diffusion models evolve toward larger DiT architectures, fine-tuning requires substantial computational resources, limiting its practical applicability.
%In contrast, training-free personalization methods\cite{personalize,wu2024infiniteid,jiang2025infiniteyou,ye2023ipadapter} eliminate the need to fine-tune model parameters and instead integrate identity features into the diffusion process through various identity injection mechanisms. These methods effectively preserve subject identity while significantly reducing computational requirements, allowing ordinary users to perform personalized image generation with consumer-grade computing devices. These advances have further broadened the applicability of personalized diffusion models.

Recent advances in text-to-image (T2I) diffusion models \cite{LDM,2} have spurred growing interest in personalized image generation, which aims to synthesize high-fidelity images of specific subjects from only a few reference images \cite{blip,face0,rb}. Existing personalization methods can be broadly divided into two categories: \textbf{training-based} and \textbf{training-free} approaches. Training-based methods \cite{textualinversion,dreambooth,lora,custom} adapt pretrained diffusion models by learning subject-specific representations or fine-tuning model parameters. Despite their effectiveness, these methods typically require additional optimization for each subject, incurring considerable computational and storage overhead. In contrast, training-free methods \cite{personalize,wu2024infiniteid,jiang2025infiniteyou,ye2023ipadapter,pulid,freecustom,photomaker} introduce subject-specific information into the generation process through dedicated conditioning mechanisms, eliminating the need for subject-specific model fine-tuning. By reducing the computational overhead associated with personalization while providing explicit identity conditioning, these methods offer a more efficient and accessible alternative for personalized image generation.
\begin{figure}
    \centering
    \includegraphics[width=1\linewidth]{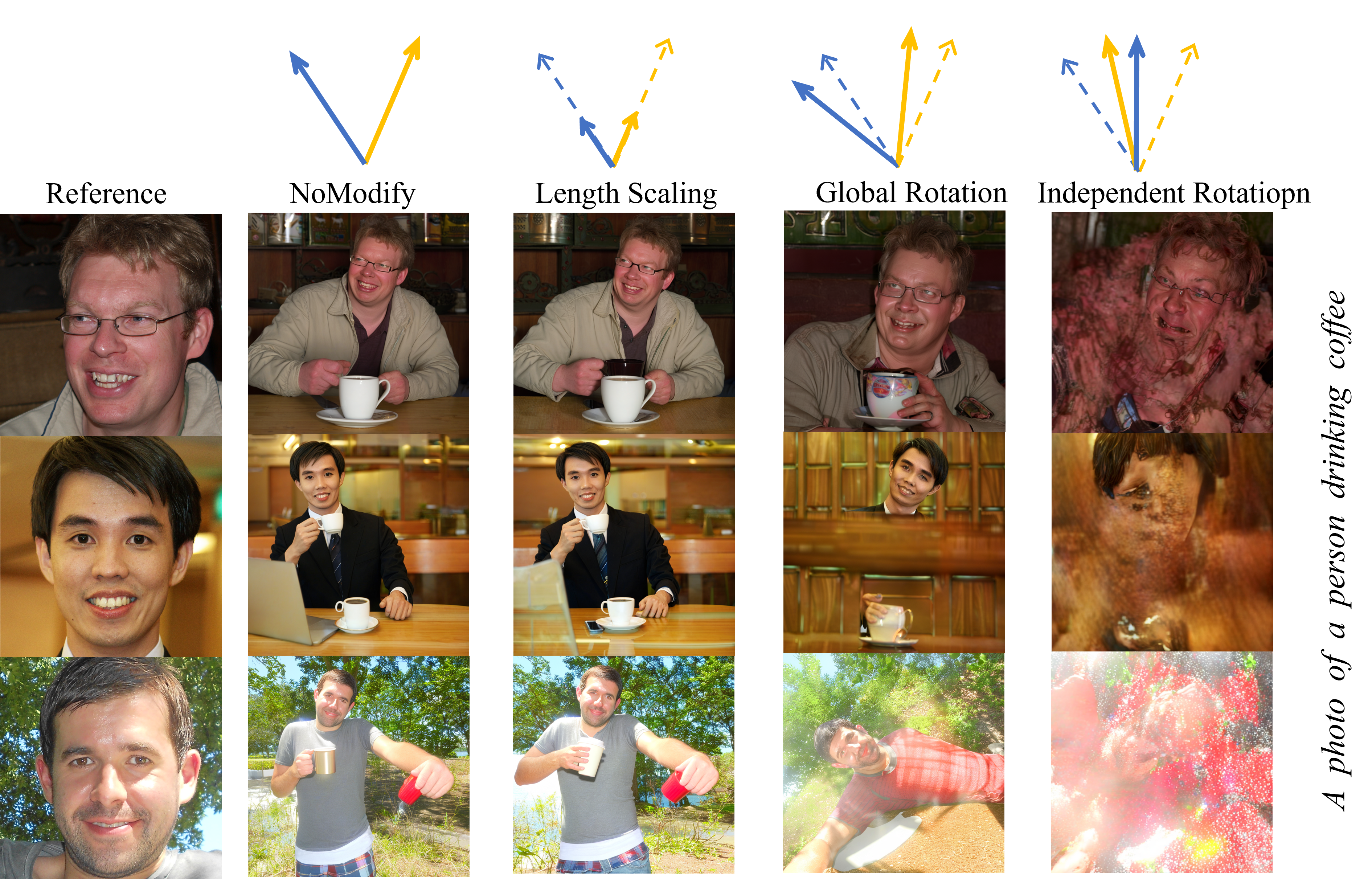}
    \caption{Effects of different geometric transformations applied to cross-attention Query (Q) features on identity preservation in personalized diffusion models. The vector diagrams illustrate the corresponding transformations. From left to right, the columns show the reference images and generated results obtained using the original Q features, length scaling, shared orthogonal rotation, and independent orthogonal rotations, respectively.}
    \label{fig:fig1}
\end{figure}

 However, the ability of personalized diffusion models to capture and reproduce subject-specific identity features also raises significant privacy concerns. Given only a few publicly available facial images, attackers can personalize diffusion models to generate unauthorized images that preserve an individual's identity, potentially violating personal privacy and portrait rights. To mitigate these risks, proactive defenses \cite{wang2025facelock,ace,decontext,antidreambooth,photoguard,pretender} have been developed to prevent unauthorized identity replication by perturbing reference images before they are used for personalization. Existing approaches target different components of the personalization pipeline. Some perturb latent representations \cite{pca,void} or interfere with image encoders \cite{idprotector} to disrupt identity feature extraction, while others \cite{idguardian} manipulate identity-conditioning signals or alter the denoising process to weaken identity-guided generation. 
 Despite these efforts, our experiments reveal that personalized models can still generate images with high identity similarity from protected reference images, particularly when the target models differ from the surrogate models used to optimize the perturbations. This limitation raises an important question: Why do perturbations targeting different stages of personalization fail to consistently prevent identity replication? We hypothesize that existing defenses may overlook the structural relationships among identity-related features. Although these methods modify feature representations or denoising behavior, such changes do not necessarily disrupt the underlying geometric structure through which identity information is integrated and preserved.

To investigate this hypothesis, we focus on the geometric structure of cross-attention Query (Q) features in personalized diffusion models. In attention-based personalization, Q features are derived from evolving image representations and determine how image tokens interact with identity-conditioning features through cross-attention. Their geometric organization may therefore influence how identity information is integrated during generation. This motivates us to investigate whether identity preservation is more sensitive to changes in Q-feature magnitudes or disruptions to their relative geometric relationships \cite{oce}.
As illustrated in Figure \ref{fig:fig1}, we conduct three controlled interventions on Q features: (i) modifying feature norms while retaining their directions; (ii) applying a shared orthogonal rotation, which changes global orientation while preserving pairwise angular relationships; and (iii) applying independent orthogonal rotations, which preserve individual norms but disrupt pairwise angular relationships. The results show that the first two interventions largely preserve identity consistency, whereas independent rotations substantially weaken identity preservation. These observations suggest that identity preservation is particularly sensitive to disruptions in the relative geometry of Q features, highlighting their potential importance in identity-conditioned generation.

%We further examine how existing proactive defenses affect the model's internal feature space. As shown in Figure\ref{fig:fig2}, we compare three representative methods, IDGuardian\cite{idguardian}, VOID\cite{void}, and IDProtector\cite{idprotector}, with our proposed method, RotGuard, focusing on their effects on the pairwise angular relationships among feature vectors in Q. In our experiments, perturbations generated by IDGuardian, VOID, and IDProtector cause relatively small changes in these relationships, whereas RotGuard substantially alters the original angular structure. This observation offers a possible explanation for the limited protection provided by existing defenses from the perspective of internal feature representations. Although these methods introduce perturbations in the input space or intermediate feature representations, they do not explicitly optimize against the relative geometry among identity feature vectors and therefore may leave internal identity representations insufficiently disrupted.

We further examine three representative proactive defenses, including IDGuardian \cite{idguardian}, VOID \cite{void}, and IDProtector \cite{idprotector}, to assess their effects on the relative geometry of Q features. As shown in Figure \ref{fig:fig2}, these methods induce only limited changes in the pairwise angular relationships among Q feature vectors. This observation suggests that existing defenses may leave much of the underlying Query geometry intact, despite perturbing different components of the personalization pipeline. Together with our controlled intervention results, this finding motivates a geometry-guided defense that explicitly targets these structural relationships.

Motivated by these findings, we propose \textbf{RotGuard}, a geometry-guided framework for transferable identity protection in personalized diffusion models. RotGuard introduces two complementary rotation-based components: \textbf{\textit{i) Query Structure Disruption (QSD)}}, which disrupts the relative angular relationships among cross-attention Query features to interfere with identity feature fusion; and \textbf{\textit{ii) Conditional Denoising Shift (CDS)}}, which extends geometric disruption to identity-guidance vectors, altering their angular relationships to redirect identity-conditioned denoising trajectories. These components are complemented by \textbf{\textit{iii) Identity Feature Space Repulsion (IFSR)}}, which weakens identity feature extraction by perturbing representations in image and identity feature spaces. By jointly optimizing these objectives on a surrogate model, RotGuard learns input-space perturbations designed to disrupt identity preservation across different personalization architectures without modifying the target models. Extensive experiments on U-Net- and DiT-based models demonstrate its effectiveness, cross-architecture transferability, and robustness against post-processing and purification attacks. Our main contributions are summarized as follows:

\begin{figure}
    \centering
    \includegraphics[width=1\linewidth]{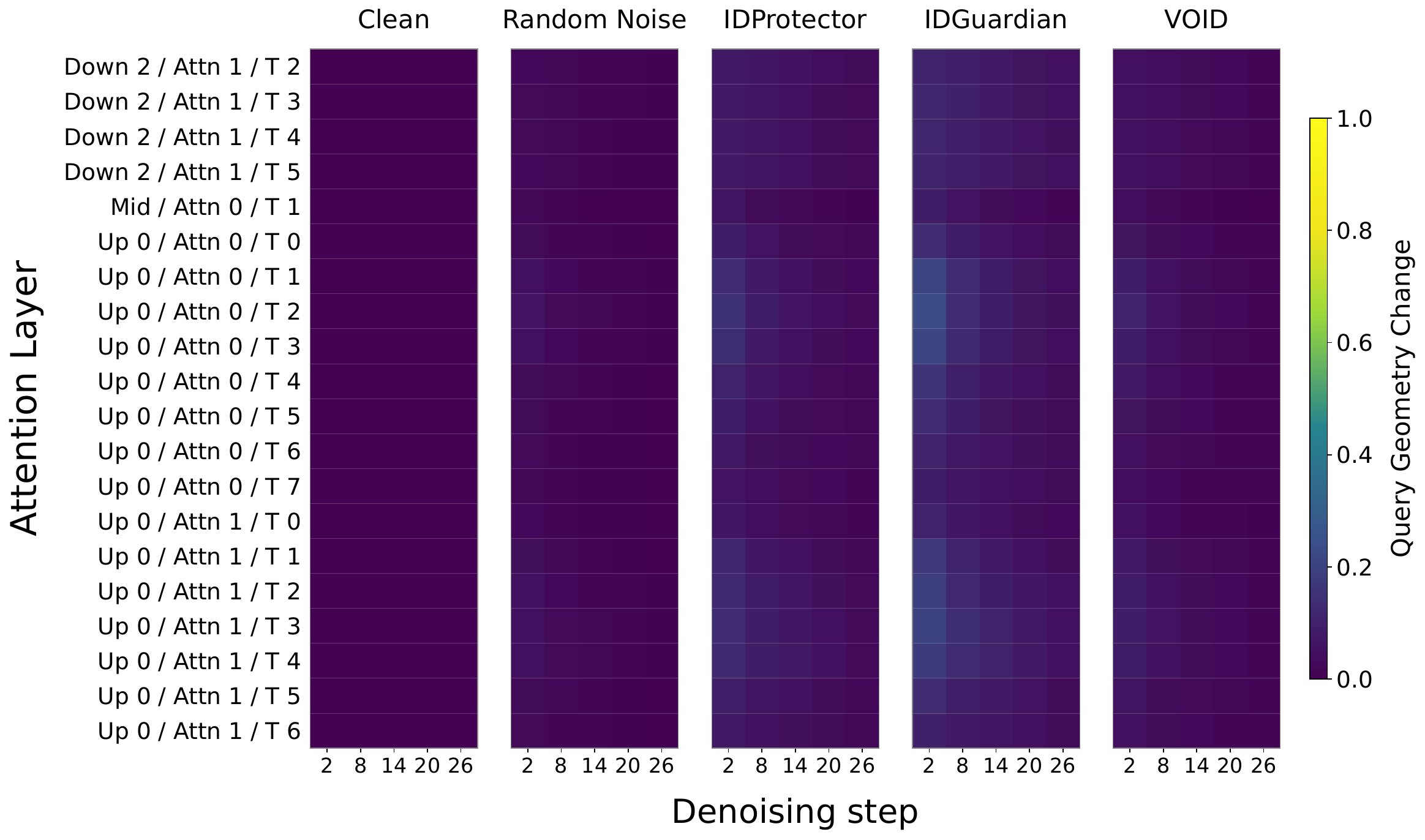}
    \caption{Geometric changes in Query features induced by different protection methods. The horizontal axis represents the denoising timestep, while the vertical axis indicates the U-Net locations from which Query features are extracted. Down, Mid, and Up denote the downsampling, middle, and upsampling stages, respectively. Attn and T represent the indices of the attention module and its internal Transformer block, respectively. All indices are zero-based.
}
    \label{fig:fig2}
\end{figure}
\begin{itemize}
    \item We investigate the role of internal feature geometry in identity preservation within personalized diffusion models. Through controlled interventions on cross-attention Query features, we reveal that disrupting their relative angular relationships is more detrimental to identity preservation than modifying feature magnitudes or global orientations.

    \item We propose \textbf{RotGuard}, a geometry-guided framework for transferable identity protection in personalized diffusion models. Building on rotation-based geometric perturbations, RotGuard introduces two complementary components: \textbf{Query Structure Disruption (QSD)}, which disrupts the relative angular relationships among cross-attention Query features to impair identity feature fusion, and \textbf{Conditional Denoising Shift (CDS)}, which alters the angular relationships among identity-guidance vectors to redirect identity-conditioned denoising trajectories.

    \item Extensive experiments demonstrate that RotGuard achieves effective and transferable identity protection across different personalization architectures, including U-Net and DiT. It also exhibits robustness against various post-processing and purification attacks, with ablation studies further validating the contributions of individual components.
\end{itemize}

\section{Related Work}
\label{sec:formatting}
\subsection{Personalized Image Generation}
With the continuous advancement of diffusion models, personalized image generation \cite{pog,instantbooth} has made significant progress. Existing personalization methods can be broadly categorized into two types: training-based and training-free approaches. Early methods primarily rely on optimizing model parameters or textual representations using a small number of reference images to achieve identity-preserving generation. Textual Inversion \cite{textualinversion} learns a new pseudo-word embedding to map a target identity into the text embedding space. DreamBooth \cite{dreambooth} further fine-tunes the diffusion model to associate a target subject with a unique identifier token, enabling identity-consistent generation across diverse scenarios. LoRA \cite{lora} further reduces the computational overhead of model fine-tuning through low-rank adaptation. Despite their promising personalization capabilities, these methods still require subject-specific optimization. Moreover, as diffusion architectures increasingly adopt larger backbones such as Diffusion Transformers (DiT) \cite{dit}, the computational cost of fine-tuning continues to increase, limiting their scalability in large-scale personalization scenarios.

To improve the efficiency and flexibility of personalized generation, subsequent studies have explored training-free approaches. IP-Adapter \cite{ye2023ipadapter} introduces decoupled cross-attention layers to inject image features extracted by a pretrained vision encoder into diffusion models. MS-Diffusion \cite{msdiffusion} employs grounding tokens and a feature resampler to enable zero-shot multi-subject personalization. Recent studies have also explored identity injection mechanisms specifically designed for DiT architectures InfiniteYou \cite{jiang2025infiniteyou} introduces InfuseNet to inject identity features into DiT through residual connections, enabling high-fidelity personalized generation. Personalize Anything \cite{personalize} achieves personalization through timestep-adaptive token replacement and patch perturbation. Furthermore, IP-Adapter-based image-conditioning mechanisms have been extended to Stable Diffusion 3.5 \cite{sd35ipa}, providing additional approaches for incorporating reference-image information into DiT-based models.
\subsection{Proactive Defense Against Unauthorized Personalization }
The rapid development of personalized diffusion models has raised concerns regarding privacy leakage and the unauthorized use of personal images. To address these concerns, proactive defense methods perturb input images to prevent personalized models from learning and reproducing identities. ACE \cite{ace} employs a predefined target to induce consistent score-function errors, thereby degrading the model's personalization capability. PCA \cite{pca} exploits the posterior collapse phenomenon in variational autoencoders (VAEs) to disrupt latent representations, while VOID \cite{void} amplifies latent encoding errors and counteracts target guidance signals by exploiting the intrinsic stochasticity of latent diffusion models, thereby suppressing identity recovery during generation. IDProtector \cite{idprotector} introduces a learned adversarial noise encoder that generates protective perturbations in a single pass. IDGuardian \cite{idguardian} further proposes a unified defense framework against both training-based and training-free personalization methods. 

Although these methods provide a certain degree of identity protection, our experiments show that personalized models can still learn identity features from protected images and reproduce the corresponding identities. Our analysis suggests that this limitation stems from existing proactive defenses primarily perturbing identity-related representations without sufficiently disrupting their internal geometric structures. The preservation of these geometric structures plays a critical role in the model's ability to maintain identity consistency during personalization. This highlights the necessity of further investigating geometry-guided identity protection methods.

\section{Threat model}
We consider an unauthorized identity personalization scenario involving two parties: an attacker and a defender.

\noindent\textbf{Attacker:} The attacker collects publicly available facial images of a target individual from online platforms and uses them as reference images in personalized diffusion models to generate novel images that preserve the target's identity. These generated images may be exploited for unauthorized identity impersonation or other malicious purposes. The attacker can employ different personalization techniques to achieve identity customization, without being restricted to a specific model architecture.

\noindent\textbf{Defender:} The defender owns the facial images and aims to prevent their unauthorized use for identity customization. Before publication, the defender adds imperceptible perturbations to images to prevent their unauthorized use for identity-preserving generation.
Given an original facial image \(x\), the defender generates a protected image \(x_p = x + \delta\), where \(\delta\) denotes the added perturbation. Let \(G(\cdot)\) represent an arbitrary personalized generator that may be employed by the attacker. The defender aims to find a perturbation \(\delta\) that minimizes the identity similarity between the original image and the generated output, while constraining the perturbation magnitude to preserve the visual quality of the protected image. This objective can be formulated as:

\[
\begin{aligned}
\arg\min_{\delta} \quad
& S_{\mathrm{id}}(x,G(x+\delta))\\
\text{s.t.}\quad
& \|\delta\|_{\infty}<\epsilon,
\end{aligned}
\]
where$S_{\mathrm{id}}(\cdot,\cdot)$ measures the identity similarity between two images, and $\epsilon$ denotes the perturbation budget, which limits the maximum magnitude of the added perturbation.

\section{Methods}
\subsection{Overall Method}
\begin{figure*}
    \centering
    \includegraphics[width=0.78\linewidth]{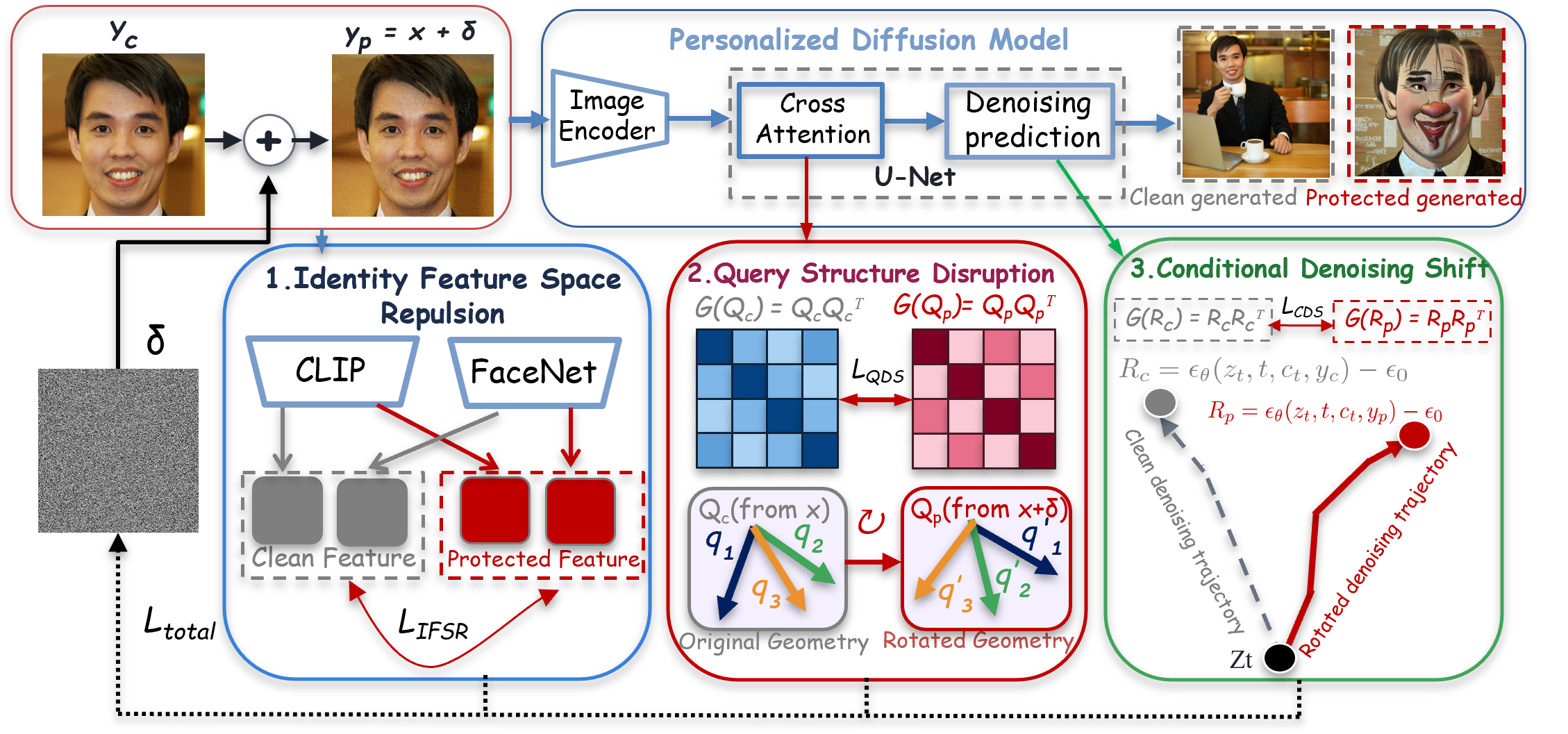}
    \caption{\textbf{Overall schematics of RotGuard.} RotGuard jointly optimizes a protective perturbation $\delta$ through three complementary components. IFSR reduces the similarity between the original and protected images in CLIP and FaceNet feature spaces. QSD disrupts the relative geometric relationships among Query features in cross-attention layers, thereby interfering with identity feature integration within the model. CDS disrupts the spatial geometric structure of identity-conditioned responses, thereby shifting the model's denoising trajectory.}
    \label{fig:method}
\end{figure*}

Our analysis in Sec.\ref{sec:intro} reveals that simply perturbing identity-related representations is insufficient to effectively undermine identity preservation in personalized diffusion models. In contrast, disrupting the internal geometric relationships among identity features can more effectively interfere with how models organize and utilize identity information. Motivated by this observation, we propose \textbf{RotGuard}, a geometry-guided identity protection framework, as illustrated in Figure \ref{fig:method}. The key idea is to jointly interfere with the learning and reproduction of target identities at three levels: identity feature extraction, identity feature integration, and conditional denoising behavior.

First, the Identity Feature Space Repulsion (IFSR) module employs CLIP \cite{clip} and FaceNet \cite{facenet} to extract visual-semantic and facial identity features, respectively. By reducing the similarity between the original and protected images in both feature spaces, IFSR weakens the model's ability to extract identity information from the input image. Second, the Query Structure Disruption (QSD) module targets Query features in cross-attention layers and employs a Gram-based objective to disrupt the relative geometric relationships among Query vectors, thereby interfering with the integration of identity features within the model. Finally, the Conditional Denoising Shift (CDS) module extracts the response vectors induced by identity conditioning in the denoising predictions and disrupts their geometric structure to suppress the influence of identity conditioning during the denoising process.
The overall optimization objective is formulated as:
\begin{equation}
\label{eq:total_loss}
\begin{gathered}
\max_{\delta}
\mathcal{L}_{\mathrm{total}}
=
\lambda_{1}\mathcal{L}_{\mathrm{IFSR}}
+
\lambda_{2}\mathcal{L}_{\mathrm{QSD}}
+
\lambda_{3}\mathcal{L}_{\mathrm{CDS}},\\
\mathrm{s.t.}\quad
\|\delta\|_\infty\leq\epsilon,
\end{gathered}
\end{equation}
where $\lambda_1$, $\lambda_2$, and $\lambda_3$ denote the weights of the three loss terms, respectively, and $\epsilon$ controls the perturbation budget.
\subsection{Identity Feature Space Repulsion}
Personalized diffusion models typically rely on pretrained visual encoders to extract identity-related features from reference images and use them as conditioning information for subsequent generation. To suppress the model's identity extraction capability, we introduce the Identity Feature Space Repulsion (IFSR) module, which jointly interferes with identity representations in complementary visual-semantic and facial identity feature spaces.

Specifically, we employ a CLIP image encoder and a FaceNet face recognition encoder to extract identity features from the input images. Given an encoder $E_k(\cdot)$, where $k\in\{\mathrm{CLIP},\mathrm{FaceNet}\}$, the normalized feature representations of the original image $x$ and the protected image $x_p$ are defined as:

\begin{equation}
f_c^k=
\frac{E_k(x)}{\|E_k(x)\|_2},
\qquad
f_p^k=
\frac{E_k(x_p)}{\|E_k(x_p)\|_2}.
\end{equation}

To reduce the similarity between the protected and original images across different feature spaces, we formulate a feature repulsion loss based on cosine distance:
\begin{equation}
\mathcal{L}_{\mathrm{IFSR}}
=
\sum_{k\in\{\mathrm{CLIP},\mathrm{FaceNet}\}}
\left(1-(f_p^k)^\top f_c^k\right),
\end{equation}
by maximizing this objective, the feature representations of the protected image are encouraged to deviate from those of the original image in both CLIP and FaceNet feature spaces, thereby reducing the ability of personalized models to extract identity information from reference images.

\subsection{Query Structure Disruption}

Cross-attention is an important component through which personalized diffusion models incorporate identity information to guide image generation. Specifically, Query features are obtained by linearly projecting the hidden representations of the current denoising network. Although Query features are not directly derived from reference facial images, they can be influenced by preceding feature interactions during the network computation involving identity conditioning. Therefore, disrupting the internal geometric structure of Query features can interfere with how the model organizes and utilizes identity-related information.

Let the Query output of the $l$-th cross-attention layer be:

\begin{equation}
Q^l\in\mathbb{R}^{N\times d},
\end{equation}
where $N$ denotes the number of spatial tokens and $d$ represents the feature dimension. We denote the Query features corresponding to the original and protected images as $Q_c^l$ and $Q_p^l$, respectively. We first apply $L_2$ normalization to each feature vector corresponding to a spatial token:

\begin{equation}
\hat{Q}_i^l
=
\frac{Q_i^l}{\|Q_i^l\|_2},
\end{equation}
then construct a Gram matrix from the normalized Query features to characterize the pairwise similarities among feature vectors at different spatial positions:

\begin{equation}
G(Q^l)=\hat{Q}^l(\hat{Q}^l)^\top,
\end{equation}
each element of the Gram matrix represents the cosine similarity between the feature vectors of two spatial positions. Therefore, it captures the relative angular relationships among Query features across different spatial positions, independently of their original magnitudes.

Based on this property, we introduce a Query Gram loss that maximizes the discrepancy between the Gram matrices of the original and protected Query representations, thereby disrupting their original geometric structure:
\begin{equation}
\mathcal{L}_{\mathrm{QSD}}
=
\frac{1}{L}\sum_{l=1}^{L}
\left\|
G(Q_p^l)-G(Q_c^l)
\right\|_F^2,
\end{equation}
where $L$ denotes the number of selected cross-attention layers and $\|\cdot\|_F$ represents the Frobenius norm.

Since the normalized Gram matrix is invariant to individual changes in feature magnitudes, QSD directly targets the geometric relationships among feature vectors. By encouraging the Query representations induced by the protected image to deviate from the original geometric structure, QSD interferes with the integration of identity information within the model.

\subsection{Conditional Denoising Shift}

Personalized diffusion models not only need to extract and integrate identity features but also rely on conditional denoising networks to translate these features into identity-consistent generation results. Therefore, we further introduce the Conditional Denoising Shift (CDS) module to interfere with the influence of identity conditioning on denoising predictions.

Given a diffusion timestep $t$, a noisy latent $z_t$, and a text condition $c_t$, we obtain denoising predictions conditioned on the identity information extracted from the original and protected images, respectively. Let $y_c$ and $y_p$ denote the corresponding identity conditions, and let the denoising network be represented by $\epsilon_\theta(\cdot)$. To isolate the contribution of identity conditioning, we first compute a baseline prediction by setting the identity image tokens to zero:

\begin{equation}
\epsilon_0
=
\epsilon_\theta(z_t,t,c_t,\mathbf{0}),
\end{equation}
where $\mathbf{0}$ denotes the identity conditioning input with all image tokens set to zero. Subsequently, we calculate the identity-conditioned responses $R_c$ and $R_p$ induced by the original and protected images, respectively:

\begin{equation}
R_c=
\epsilon_\theta(z_t,t,c_t,y_c)-\epsilon_0,
\end{equation}

\begin{equation}
R_p=
\epsilon_\theta(z_t,t,c_t,y_p)-\epsilon_0.
\end{equation}

These responses characterize the incremental effects of identity conditioning on denoising predictions relative to the identity-free baseline. Unlike directly comparing two complete denoising predictions, this differential formulation isolates the response induced by identity conditioning while keeping the text condition and other shared factors fixed.

We then normalize $R_c$ and $R_p$ and construct their corresponding Gram matrices. Specifically, the response features are reshaped into spatial feature representations, and each feature vector is normalized using $L_2$ normalization:

\begin{equation}
\hat{R}_i=
\frac{R_i}{\|R_i\|_2},
\qquad i\in\{c,p\},
\end{equation}
%The corresponding spatial Gram matrix is defined as:
%\begin{equation}
%G(R)=\hat{R}\hat{R}^{\top},
%\end{equation}
based on these representations, we define the conditional denoising shift loss as:
\begin{equation}
\mathcal{L}_{\mathrm{CDS}}
=
\frac{1}{N^2}
\left\|
G(R_p)-G(R_c)
\right\|_F^2.
\end{equation}

By maximizing the discrepancy between the Gram matrices of the identity-conditioned responses induced by the original and protected images, CDS alters the geometric relationships among identity-guidance responses. Since these responses directly contribute to denoising predictions, this constraint further interferes with the influence of identity conditioning on the generation process, making it difficult for the model to maintain the original identity-conditioned denoising response structure.

\section{Experiments}

\begin{table*}[t]
    \centering
    \caption{Comparison of different identity protection methods. ISM is the average identity similarity (ArcFace, FaceNet, VGGFace), ID-Sim measures visual identity consistency between images. IP-XL denotes IP-Adapter SDXL Plus Face, SD3.5-IP denotes Stable Diffusion 3.5 Large IP-Adapter, Per-Any denotes Personalize-Anything, DeC denotes DeContext, IDP denotes IDProtector, IDG denotes IDGuardian. The best results are highlighted in bold.}
    \label{tab:comparison}
    \small
    \setlength{\tabcolsep}{0.2pt}
    \renewcommand{\arraystretch}{1.25}
    \resizebox{\textwidth}{!}{
    \begin{tabular*}{\textwidth}{@{\extracolsep{\fill}}l*{18}{c}@{}}
        \toprule
        \multirow{2}{*}{\shortstack{\textbf{Method}}}
& \multicolumn{2}{c}{\footnotesize\textbf{\shortstack{IP-Adapter}}}
& \multicolumn{2}{c}{\footnotesize\textbf{\shortstack{IP-XL}}}
& \multicolumn{2}{c}{\footnotesize\textbf{InfiniteYou}}
& \multicolumn{2}{c}{\footnotesize\textbf{SD3.5-IP}}
& \multicolumn{2}{c}{\footnotesize\textbf{MUSE}}
& \multicolumn{2}{c}{\footnotesize\textbf{InstantID}}
& \multicolumn{2}{c}{\footnotesize\textbf{\shortstack{MS-Difussion}}}
& \multicolumn{2}{c}{\footnotesize\textbf{Dreambooth}}
& \multicolumn{2}{c}{\footnotesize\textbf{\shortstack{Per-Any}}} \\

        \cmidrule(lr){2-3}
        \cmidrule(lr){4-5}
        \cmidrule(lr){6-7}
        \cmidrule(lr){8-9}
        \cmidrule(lr){10-11}
        \cmidrule(lr){12-13}
        \cmidrule(lr){14-15}
        \cmidrule(lr){16-17}
        \cmidrule(lr){18-19}

& {\footnotesize ISM$\downarrow$} & {\footnotesize IDSIM$\downarrow$}
& {\footnotesize ISM$\downarrow$} & {\footnotesize IDSIM$\downarrow$}
& {\footnotesize ISM$\downarrow$} & {\footnotesize IDSIM$\downarrow$}
& {\footnotesize ISM$\downarrow$} & {\footnotesize IDSIM$\downarrow$}
& {\footnotesize ISM$\downarrow$} & {\footnotesize IDSIM$\downarrow$}
& {\footnotesize ISM$\downarrow$} & {\footnotesize IDSIM$\downarrow$}
& {\footnotesize ISM$\downarrow$} & {\footnotesize IDSIM$\downarrow$}
& {\footnotesize ISM$\downarrow$} & {\footnotesize IDSIM$\downarrow$}
& {\footnotesize ISM$\downarrow$} & {\footnotesize IDSIM$\downarrow$} \\

        \midrule

        \textbf{No Protected}
        & 0.564 & 0.502
        & 0.608 & 0.503
        & 0.659 & 0.357
        & 0.375 & 0.390
        & 0.478 & 0.639
        & 0.600 & 0.333
        & 0.471 & 0.448
        & 0.636 & 0.660
        & 0.504 & 0.357 \\

        \textbf{DeC} \cite{decontext}
        & 0.432 & 0.444
        & 0.495 & 0.449
        & 0.592 & 0.361
        & 0.307 & 0.365
        & 0.441 & 0.597
        & 0.354 & 0.245
        & 0.426 & 0.483
        & 0.557 & 0.621
        & 0.419 & 0.323 \\

        \textbf{VOID} \cite{void}
        & 0.506 & 0.493
        & 0.544 & 0.498
        & 0.639 & 0.353
        & 0.325 & 0.362
        & 0.459 & 0.639
        & 0.438 & 0.253
        & 0.404 & \textbf{0.461}
        & 0.609 & 0.672
        & 0.451 & 0.330 \\

        \textbf{PCA} \cite{pca}
        & 0.496 & 0.479
        & 0.547 & 0.483
        & 0.638 & 0.355
        & 0.332 & 0.355
        & 0.453 & 0.633
        & 0.414 & 0.257
        & 0.413 & 0.467
        & 0.617 & 0.665
        & 0.401 & 0.303 \\

        \textbf{ACE} \cite{ace}
        & 0.529 & 0.495
        & 0.570 & 0.496
        & 0.628 & 0.354
        & 0.337 & 0.371
        & 0.468 & 0.641
        & 0.429 & 0.256
        & 0.408 & 0.472
        & 0.594 & 0.655
        & 0.467 & 0.338 \\

        \textbf{IDP} \cite{idprotector}
        & 0.486 & 0.461
        & 0.513 & 0.510
        & 0.576 & 0.347
        & 0.316 & 0.364
        & 0.406 & 0.556
        & 0.531 & 0.304
        & 0.407 & 0.471
        & 0.606 & 0.633
        & 0.445 & 0.337 \\

        \textbf{IDG} \cite{idguardian}
        & 0.360 & 0.394
        & 0.386 & 0.398
        & 0.533 & 0.351
        & 0.298 & 0.354
        & 0.394 & 0.601
        & \textbf{0.160} & 0.166
        & 0.418 & 0.482
        & \textbf{0.541} & 0.631
        & 0.478 & 0.330 \\

        \textbf{Ours}
        & \textbf{0.244} & \textbf{0.286}
        & \textbf{0.253} & \textbf{0.242}
        & \textbf{0.503} & \textbf{0.305}
        & \textbf{0.258} & \textbf{0.299}
        & \textbf{0.291} & \textbf{0.446}
        & 0.166 & \textbf{0.164}
        & \textbf{0.353} & 0.480
        & 0.574 & \textbf{0.595}
        & \textbf{0.398} & \textbf{0.281} \\

        \bottomrule
    \end{tabular*}
    }
\end{table*}

\begin{figure*}[t]
    \centering
    \includegraphics[width=0.79\linewidth]{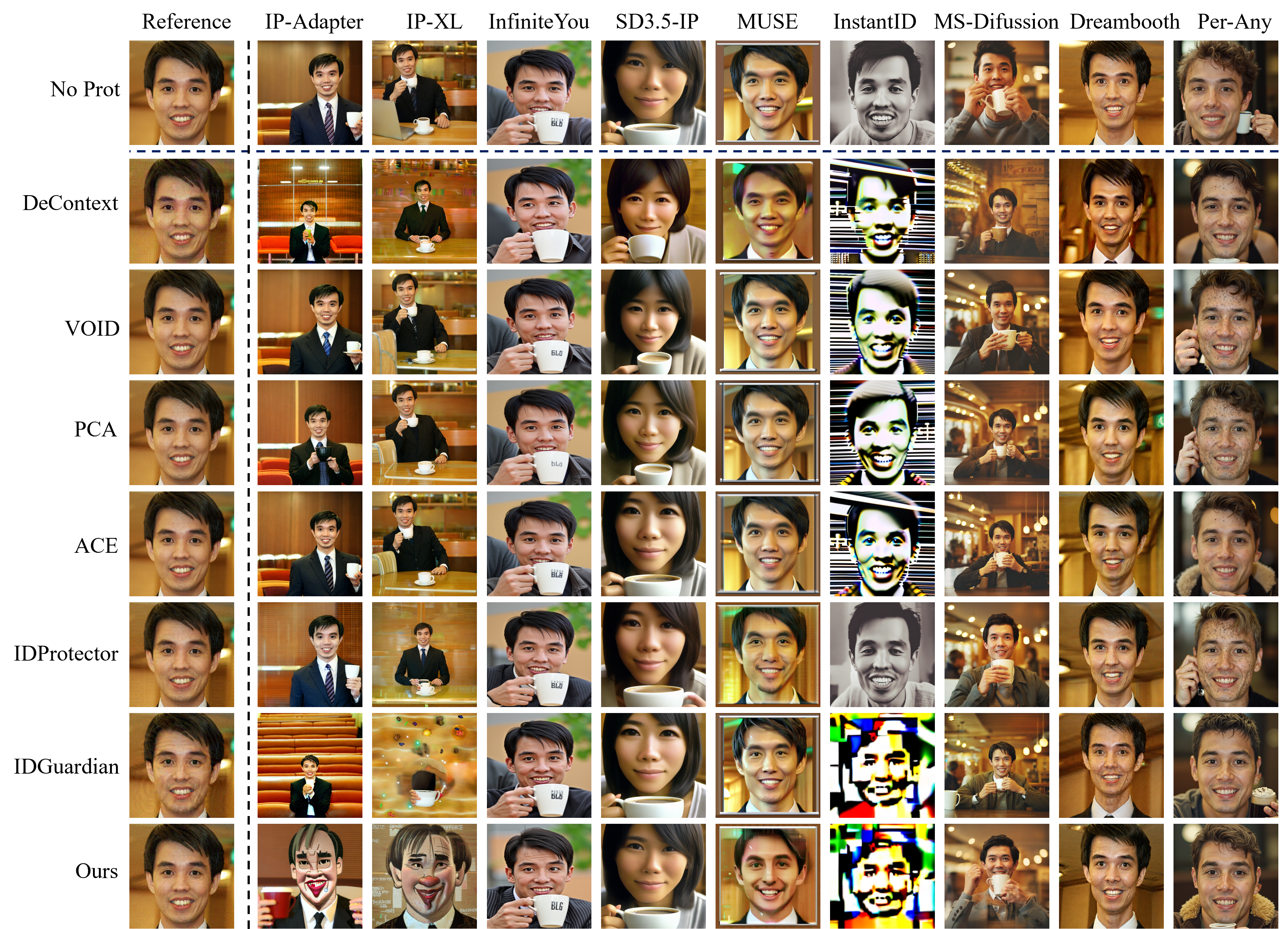}
    \caption{Qualitative comparison of different identity protection methods across diverse personalization models. IP-XL denotes IP-Adapter SDXL Plus Face, and Per-Any denotes Personalize-
Anything.}
    \label{fig:fig4}
\end{figure*}

\begin{table*}[t]
    \centering
    \caption{Robustness evaluation of RotGuard against image post-processing and purification attacks, including JPEG compression, Gaussian noise, Gaussian blur, and UDAP.}
    \label{tab:robustness}
    \small
    \setlength{\tabcolsep}{2pt}
    \renewcommand{\arraystretch}{1.2}

    \resizebox{\textwidth}{!}{%
    \begin{tabular}{l*{18}{c}}
        \toprule
        \multirow{2}{*}{\textbf{Method}}
        & \multicolumn{2}{c}{\textbf{IP-Adapter}}
        & \multicolumn{2}{c}{\textbf{IP-XL}}
        & \multicolumn{2}{c}{\textbf{InfiniteYou}}
        & \multicolumn{2}{c}{\textbf{SD3.5-IP}}
        & \multicolumn{2}{c}{\textbf{MUSE}}
        & \multicolumn{2}{c}{\textbf{InstantID}}
        & \multicolumn{2}{c}{\textbf{MS-Diffusion}}
        & \multicolumn{2}{c}{\textbf{Dreambooth}}
        & \multicolumn{2}{c}{\textbf{Per-Any}} \\

        \cmidrule(lr){2-3}
        \cmidrule(lr){4-5}
        \cmidrule(lr){6-7}
        \cmidrule(lr){8-9}
        \cmidrule(lr){10-11}
        \cmidrule(lr){12-13}
        \cmidrule(lr){14-15}
        \cmidrule(lr){16-17}
        \cmidrule(lr){18-19}

        & ISM$\downarrow$ & IDSIM$\downarrow$
        & ISM$\downarrow$ & IDSIM$\downarrow$
        & ISM$\downarrow$ & IDSIM$\downarrow$
        & ISM$\downarrow$ & IDSIM$\downarrow$
        & ISM$\downarrow$ & IDSIM$\downarrow$
        & ISM$\downarrow$ & IDSIM$\downarrow$
        & ISM$\downarrow$ & IDSIM$\downarrow$
        & ISM$\downarrow$ & IDSIM$\downarrow$
        & ISM$\downarrow$ & IDSIM$\downarrow$ \\

        \midrule

        \textbf{No Protected}
        & 0.564 & 0.502
        & 0.608 & 0.503
        & 0.659 & 0.357
        & 0.375 & 0.390
        & 0.478 & 0.639
        & 0.600 & 0.333
        & 0.471 & 0.448
        & 0.636 & 0.660
        & 0.504 & 0.357 \\

        \textbf{JPEG(q=50)}
        & 0.366 & 0.429
        & 0.387 & 0.425
        & 0.518 & 0.352
        & 0.308 & 0.365
        & 0.352 & 0.596
        & 0.163 & 0.168
        & 0.402 & 0.483
        & 0.513 & 0.564
        & 0.434 & 0.318 \\

        \textbf{Blur($\sigma=1.7$)}
        & 0.274 & 0.379
        & 0.350 & 0.395
        & 0.521 & 0.315
        & 0.323 & 0.358
        & 0.409 & 0.573
        & 0.169 & 0.170
        & 0.433 & 0.448
        & \textbf{0.495} & \textbf{0.536}
        & 0.431 & 0.294 \\

        \textbf{Noise($\sigma^2=900$)}
        & 0.321 & 0.443
        & 0.389 & 0.438
        & 0.544 & 0.332
        & 0.287 & 0.336
        & 0.431 & 0.557
        & \textbf{0.142} & \textbf{0.145}
        & 0.417 & 0.450
        & 0.511 & 0.627
        & 0.401 & 0.299 \\

        \textbf{UDAP} \cite{udap}
        & 0.405 & 0.390
        & 0.446 & 0.392
        & 0.512 & \textbf{0.286}
        & 0.299 & 0.340
        & 0.406 & 0.526
        & 0.157 & 0.166
        & 0.411 & \textbf{0.413}
        & 0.592 & 0.555
        & 0.433 & 0.298 \\

        \textbf{Ours}
        & \textbf{0.244} & \textbf{0.286}
        & \textbf{0.253} & \textbf{0.242}
        & \textbf{0.503} & 0.305
        & \textbf{0.258} & \textbf{0.299}
        & \textbf{0.291} & \textbf{0.446}
        & 0.166 & 0.164
        & \textbf{0.353} & 0.480
        & 0.574 & 0.595
        & \textbf{0.398} & \textbf{0.281} \\

        \bottomrule
    \end{tabular}%
    }
\end{table*}

\begin{table*}[t]
    \centering
    \caption{Ablation study of progressively incorporating IFSR, QSD, and CDS into RotGuard. "Ours" denotes IFSR+QSD+CDS.}
    \label{tab:ablation}
    \small
    \setlength{\tabcolsep}{2pt}
    \renewcommand{\arraystretch}{1.2}

    \resizebox{\textwidth}{!}{%
    \begin{tabular}{l*{18}{c}}
        \toprule
        \multirow{2}{*}{\textbf{Method}}
        & \multicolumn{2}{c}{\textbf{IP-Adapter}}
        & \multicolumn{2}{c}{\textbf{IP-XL}}
        & \multicolumn{2}{c}{\textbf{InfiniteYou}}
        & \multicolumn{2}{c}{\textbf{SD3.5-IP}}
        & \multicolumn{2}{c}{\textbf{MUSE}}
        & \multicolumn{2}{c}{\textbf{InstantID}}
        & \multicolumn{2}{c}{\textbf{MS-Diffusion}}
        & \multicolumn{2}{c}{\textbf{Dreambooth}}
        & \multicolumn{2}{c}{\textbf{Per-Any}} \\

        \cmidrule(lr){2-3}
        \cmidrule(lr){4-5}
        \cmidrule(lr){6-7}
        \cmidrule(lr){8-9}
        \cmidrule(lr){10-11}
        \cmidrule(lr){12-13}
        \cmidrule(lr){14-15}
        \cmidrule(lr){16-17}
        \cmidrule(lr){18-19}

        & ISM$\downarrow$ & IDSIM$\downarrow$
        & ISM$\downarrow$ & IDSIM$\downarrow$
        & ISM$\downarrow$ & IDSIM$\downarrow$
        & ISM$\downarrow$ & IDSIM$\downarrow$
        & ISM$\downarrow$ & IDSIM$\downarrow$
        & ISM$\downarrow$ & IDSIM$\downarrow$
        & ISM$\downarrow$ & IDSIM$\downarrow$
        & ISM$\downarrow$ & IDSIM$\downarrow$
        & ISM$\downarrow$ & IDSIM$\downarrow$ \\

        \midrule

        \textbf{No Protected}
        & 0.564 & 0.502
        & 0.608 & 0.503
        & 0.659 & 0.357
        & 0.375 & 0.390
        & 0.478 & 0.639
        & 0.600 & 0.333
        & 0.471 & 0.448
        & 0.636 & 0.660
        & 0.504 & 0.357 \\

        \textbf{IFSR}
        & 0.298 & 0.321
        & 0.341 & 0.281
        & 0.621 & 0.358
        & 0.318 & 0.352
        & 0.399 & 0.552
        & \textbf{0.149} & 0.263
        & 0.422 & 0.487
        & 0.605 & 0.601
        & 0.441 & 0.319 \\

        \textbf{IFSR+QSD}
        & 0.280 & 0.308
        & 0.273 & \textbf{0.233}
        & 0.614 & 0.330
        & 0.314 & 0.328
        & 0.419 & 0.569
        & 0.162 & 0.166
        & 0.405 & \textbf{0.478}
        & 0.592 & 0.646
        & 0.452 & 0.321 \\

        \textbf{Ours}
        & \textbf{0.244} & \textbf{0.286}
        & \textbf{0.253} & 0.242
        & \textbf{0.503} & \textbf{0.305}
        & \textbf{0.258} & \textbf{0.299}
        & \textbf{0.291} & \textbf{0.446}
        & 0.166 & \textbf{0.164}
        & \textbf{0.353} & 0.480
        & \textbf{0.574} & \textbf{0.595}
        & \textbf{0.398} & \textbf{0.281} \\

        \bottomrule
    \end{tabular}%
    }
\end{table*}
\subsection{Experimental Setup}
 % unnumbered
\noindent{\textbf{Dataset.}} We evaluate RotGuard on FFHQ \cite{ffhq}, a widely used high-resolution face dataset. To ensure both representativeness and experimental controllability, we curate a subset containing 300 distinct identities. All images are processed with a unified pipeline: we perform center cropping with scale normalization to align facial regions, and then resize all samples to a resolution of $1024\times1024$ pixels. This setup preserves the fine-grained details and identity cues in FFHQ while matching the standard input configuration of high-resolution diffusion models, enabling a fair comparison of identity protection effectiveness and visual quality across methods.

\noindent{\textbf{Training Details.}} We generate protective perturbations using IP-Adapter SDXL Plus Face \cite{ye2023ipadapter} as the surrogate personalization model, which adopts the pretrained SDXL architecture \cite{sdxl} as its backbone. The perturbation optimization is performed via Projected Gradient Descent (PGD) \cite{pgd}, where the conditioning prompt is fixed as \(c=\) "A photo of a person". We set the maximum perturbation magnitude to \(\epsilon=8/255\) and use a step size of \(\alpha=0.005\) for 100 optimization iterations. The experiments are carried out on a single NVIDIA RTX Pro 6000 GPU equipped with 96 GB of memory.

\noindent{\textbf{Evaluation Metrics.}}
We evaluate RotGuard from two perspectives: identity protection effectiveness and visual fidelity.
For identity protection, we employ two metrics: Identity Score Matching (ISM) and IDSIM \cite{idsim}. ISM measures the identity feature consistency between the original reference images and the generated images. We extract facial features from the reference and generated images using ArcFace \cite{arcface}, FaceNet \cite{facenet}, and VGG-Face \cite{vggface}, calculate the cosine similarity between the corresponding features, and average the scores obtained from the three models. Additionally, we adopt IDSIM, a learning-based identity-aware metric that evaluates visual identity consistency while maintaining robustness to contextual variations
To assess visual fidelity, we employ PSNR and SSIM to evaluate the impact of protective perturbations on the visual quality of the original images.

\subsection{Comparison With Existing Methods}
\noindent{\textbf{Protection Performance.}}
We compare RotGuard with six representative proactive defense methods \cite{void,ace,idguardian,idprotector,decontext,pca} to evaluate their identity protection capabilities across diverse personalization pipelines. We select nine personalization methods covering two backbone architectures: U-Net and DiT.
The U-Net-based personalization methods include IP-Adapter \cite{ye2023ipadapter}, IP-Adapter SDXL Plus Face \cite{ye2023ipadapter}, MUSE \cite{muse}, InstantID \cite{instantid}, MS-Diffusion \cite{msdiffusion}, and DreamBooth \cite{dreambooth}. The DiT-based methods include InfiniteYou \cite{jiang2025infiniteyou}, Stable Diffusion 3.5 Large IP-Adapter \cite{sd35ipa}, and Personalize-Anything \cite{personalize}. 

Table \ref{tab:comparison} presents the quantitative comparison results. RotGuard achieves the lowest or near-lowest scores in terms of both ISM and IDSIM, indicating its effectiveness in disrupting identity consistency preservation across diverse personalization models.In addition to quantitative comparisons, we present qualitative results in Figure \ref{fig:fig4}, which further demonstrate that RotGuard effectively suppresses identity preservation across different personalization methods and achieves better protection performance than the baseline methods.

\begin{table}[t]
    \centering
    \caption{Image quality (PSNR,SSIM) comparison of different identity protection methods. DeC denotes DeContext, IDP denotes IDProtector, IDG denotes IDGuardian.The best results are highlighted in bold, while the second-best results are underlined.}
    \label{tab:image_quality}
    \small
    \setlength{\tabcolsep}{2pt}
    \renewcommand{\arraystretch}{1.2}
    \begin{tabular}{lccccccc}
        \toprule
        \textbf{Metric}
        & \textbf{DeC}
        & \textbf{VOID}
        & \textbf{PCA}
        & \textbf{ACE}
        & \textbf{IDP}
        & \textbf{IDG}
        & \textbf{Ours} \\
        \midrule
        SSIM $\uparrow$
        & 0.696
        & \underline{0.882}
        & 0.804
        & 0.751
        & 0.793
        & 0.842
        & \textbf{0.886} \\

        PSNR $\uparrow$
        & 29.302
        & \textbf{34.456}
        & 32.685
        & 30.035
        & 28.478
        & 32.028
        & \underline{33.288} \\
        \bottomrule
    \end{tabular}
\end{table}

\noindent{\textbf{Image Quality.}} Table \ref{tab:image_quality} presents the impact of different defense methods on the visual quality of the original images. The results show that RotGuard achieves effective identity protection while maintaining good visual fidelity compared with other methods.

\subsection{Robustness Evaluation}
We evaluate the robustness of RotGuard against different types of attacks, including three common image post-processing operations: JPEG compression with a quality factor of 50, Gaussian noise with a variance of 900, and Gaussian blur with \(\sigma=1.7\), as well as UDAP \cite{udap}, a latest purification method \cite{dirp,freepure}. As shown in Table 3, RotGuard maintains effective identity protection even after relatively strong image processing operations. Notably, some post-processing and purification operations can even further improve the identity protection performance of RotGuard in certain personalization scenarios. 

\subsection{Ablation Studies}
\noindent{\textbf{The Role Of Components.}}
Table \ref{tab:ablation} presents the ablation results of different component combinations in RotGuard. Using IFSR alone reduces the ISM scores compared with unprotected images, demonstrating its effectiveness in interfering with identity feature extraction. Building upon IFSR, incorporating QSD further improves identity protection, indicating that disrupting the geometric relationships among cross-attention Query features can further weaken identity preservation. Finally, adding CDS leads to further improvements in most personalization models, demonstrating that perturbing the influence of identity conditions on the denoising trajectory provides additional protection. These results highlight the complementary contributions of the three components, whose combination enables more effective identity protection.

\noindent{\textbf{Perturbation Budget.}}
We further investigate the effects of the perturbation budget \(\epsilon\) on RotGuard's performance by comparing identity protection effectiveness and visual quality under different parameter settings. As shown in Figure 1 in the Appendix, \(\epsilon=8/255\) enables RotGuard to effectively weaken identity preservation while maintaining good visual fidelity.Detailed experimental results and analysis can be found in Appendix A.

\noindent{\textbf{Number Of Optimization Steps.}}
We investigate the effect of optimization iterations by monitoring the convergence behavior of the total loss over 1,000 iterations. As shown in Fig 2 in the Appendix, the total loss increases rapidly within the first 150 iterations and gradually stabilizes after approximately 200 iterations, thereby indicating diminishing returns from additional optimization at later stages. Notably, the loss exhibits only a slight increase between 100 and 150 iterations. Considering the limited additional improvement and the associated computational cost, we set the number of optimization iterations to 100. More detailed experimental results and analysis are provided in Appendix B.

\section{Conclusion}
In this paper, we propose\textbf{ RotGuard}, a geometry-guided identity protection framework designed to prevent unauthorized personalization in diffusion models. By integrating \textbf{Identity Feature Space Repulsion (IFSR)}, \textbf{Query Structure Disruption (QSD)}, and \textbf{Conditional Denoising Shift (CDS)}, RotGuard jointly interferes with identity feature extraction, internal geometric relationships, and identity-guided denoising trajectories. Extensive experiments demonstrate the effectiveness of RotGuard across various personalization methods and backbone architectures, including U-Net and DiT. Furthermore, RotGuard maintains robust protection performance against common image post-processing and purification attacks. We hope this work highlights the importance of disrupting internal feature geometry in personalized diffusion models and provides a new perspective on identity protection.
\section*{LLM Usage Statement}
The authors used OpenAI ChatGPT-6 to assist with language editing and translation in the Introduction and Related Work sections. In addition, GPT-6.1sol was used to assist with programming tasks, including the reproduction of existing personalization methods and proactive defense methods. The core implementation of RotGuard and the evaluation of all experimental metrics were independently developed and conducted by the authors. All AI-generated suggestions and content were carefully reviewed, verified, and revised by the authors.
{
    \small
    \bibliographystyle{ieeenat_fullname}
    \bibliography{ref/ref}
}

\clearpage
\maketitlesupplementary
\section*{A. Perturbation Budget}
\begin{figure}
    \centering
    \includegraphics[width=1\linewidth]{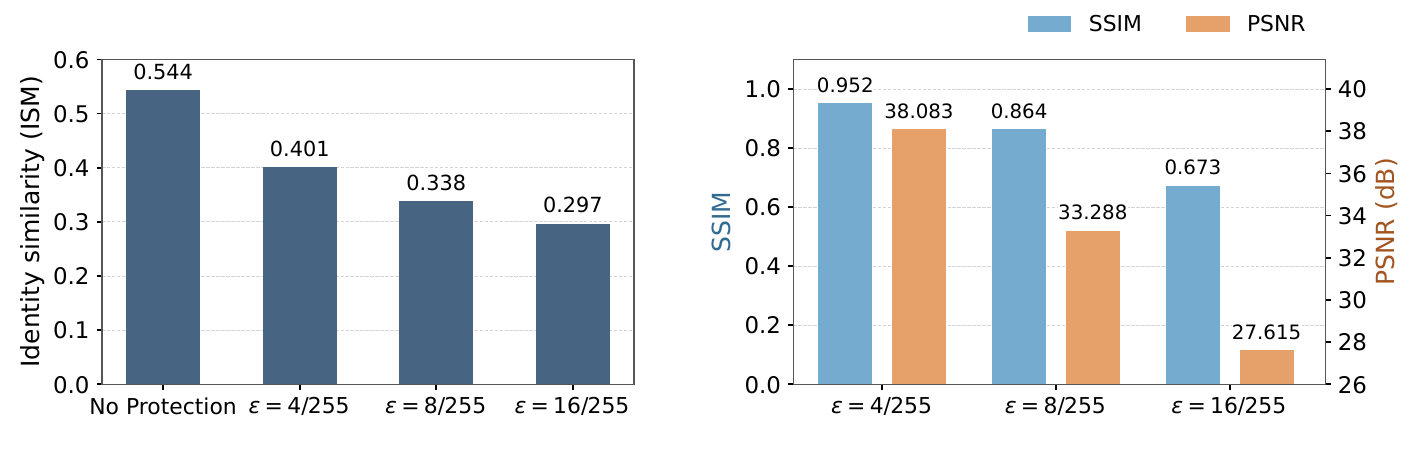}
    \caption{Ablation study of different perturbation budgets on the identity protection effectiveness and visual fidelity of RotGuard.}
    \label{fig:budget}
\end{figure}
We investigate the influence of the perturbation budget \(\epsilon\) on the identity protection effectiveness and visual fidelity of RotGuard. Specifically, we evaluate three perturbation budgets, \(\epsilon\in\{4/255,8/255,16/255\}\), while keeping all other optimization settings unchanged.

As illustrated in Figure \ref{fig:budget}, different perturbation budgets result in different trade-offs between identity protection and visual fidelity. A smaller budget (\(\epsilon=4/255\)) imposes stricter constraints on perturbation optimization, limiting the extent to which identity-related representations can be disrupted. In contrast, a larger budget (\(\epsilon=16/255\)) provides greater optimization flexibility but increases the potential for visual distortion. Considering both protection effectiveness and image quality, we select \(\epsilon=8/255\) as the default perturbation budget for RotGuard.

\section*{B. Number of Optimization Steps}
\begin{figure}
    \centering
    \includegraphics[width=1\linewidth]{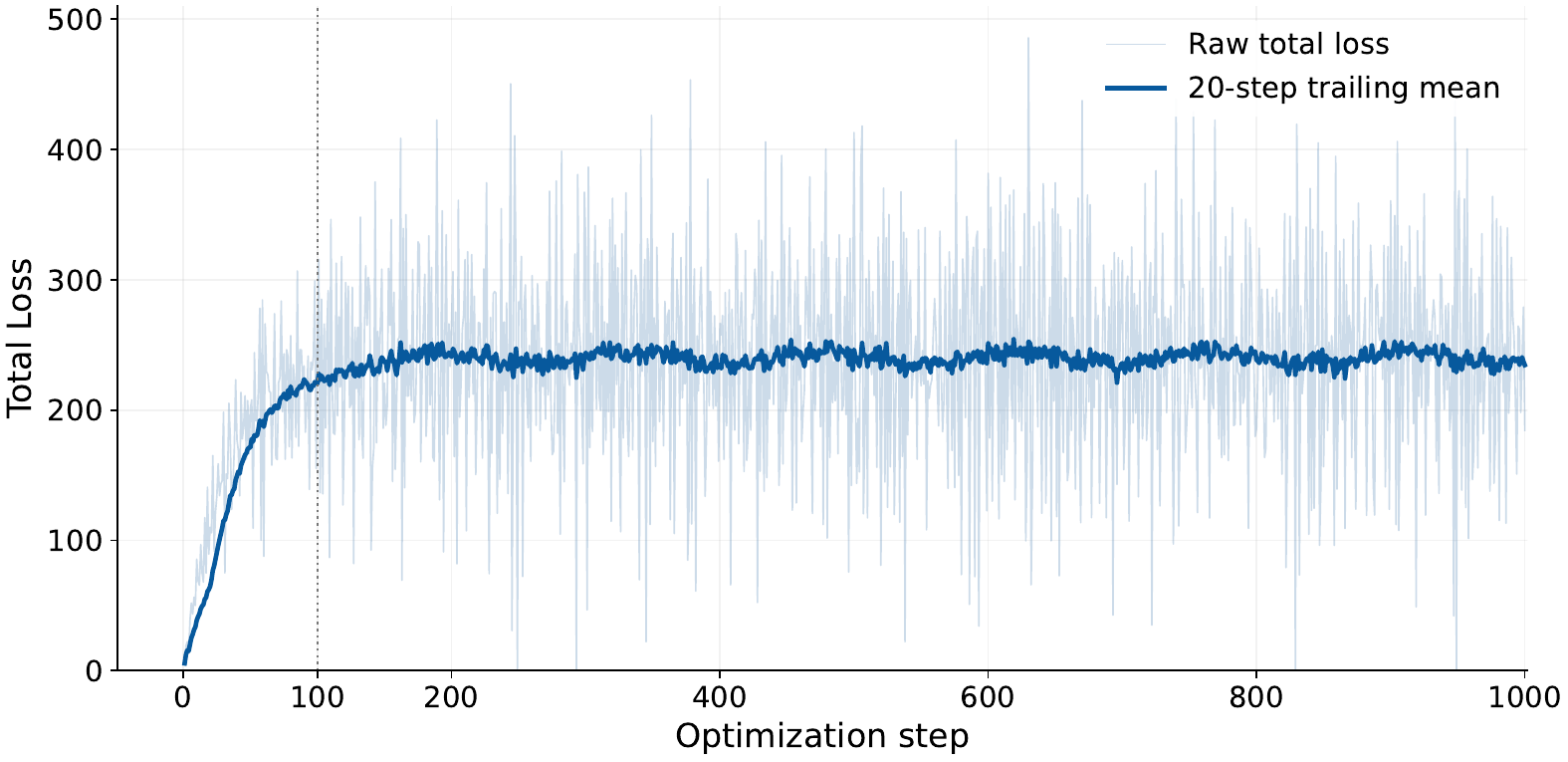}
    \caption{Convergence behavior of the total loss over 1,000 optimization iterations. The light blue curve represents the raw total loss, while the dark blue curve shows its 20-step trailing mean. It reveals rapid improvement during the early iterations, followed by gradual stabilization after approximately 200 iterations.}
    \label{fig:loss}
\end{figure}

We visualize the evolution of the total loss over 1,000 optimization iterations, as shown in Figure \ref{fig:loss}. The light blue curve represents the raw total loss at each iteration, while the dark blue curve shows its 20-step trailing mean. Specifically, each point on the dark blue curve is calculated by averaging the loss values from the current iteration and the preceding 19 iterations. For the first 19 iterations, all available values up to the current iteration are used to compute the average.
The raw total loss exhibits considerable fluctuations, primarily because the CDS and QSD losses are computed using newly sampled diffusion timesteps and noise at each iteration. Both objectives involve summing the squared differences between pairwise token similarities. Consequently, even small changes in the similarities of individual token pairs can accumulate across numerous pairs, leading to substantial variations in the total loss between consecutive iterations. To better illustrate the overall optimization trend, we employ the aforementioned moving average to smooth out these short-term fluctuations.

As indicated by the smoothed curve, the total loss increases rapidly during the early optimization stage, with the most substantial improvements occurring within the first 150 iterations. After approximately 200 iterations, the curve gradually stabilizes, indicating that further optimization yields diminishing returns. Notably, the increase in the total loss between 100 and 150 iterations is relatively small compared with that during the initial stage. Considering the limited additional improvement and the associated computational cost, we therefore select 100 iterations as a practical stopping point.

\section*{C. Method Details}
This section details the complete optimization procedure of RotGuard,
summarized in Algorithm~\ref{alg:rotguard}. The feature term is
evaluated once per update, whereas the QSD and CDS terms are estimated over
$M$ diffusion draws. We accumulate the gradients of all terms, averaging the
diffusion-based ones over the draws, before taking the sign.

\noindent\textbf{Image conditioning.}
The token extractor $T$ projects the penultimate-layer CLIP hidden states
through the IP-Adapter image projection network. For feature repulsion,
$f_e$ denotes the unit-normalized global embedding of encoder
$e\in\mathcal E=\{\mathrm{CLIP},\mathrm{FaceNet}\}$; $f_{\mathrm{CLIP}}$ is
computed with the same ViT-H backbone used by $T$. Each draw samples a
timestep $t$ and Gaussian noise, and forms the noisy latent $z_t$ with the
scheduler's cumulative coefficient $\bar\alpha_t$. The zero-token, clean, and
protected passes share the same $z_t$, $t$, and text conditioning. The clean
and zero-token outputs are treated as constants (stop-gradient), so gradients
reach $\delta$ only through the protected tokens $y_p$ and the encoder
features $f_e(x_p)$.

\noindent\textbf{Spatial Gram matrices.}
We collect each query matrix at the cross-attention query projection
(\texttt{attn2.to\_q}), before it is split into heads. A denoising response is
flattened over its spatial grid, with one row per location and one column per
channel. For either representation, let $V\in\mathbb R^{N\times C}$ be the
resulting matrix, its spatial Gram matrix is
\[
  G=\hat V\hat V^\top,\qquad
  \hat V=V/\|V\|_2.
\]

\begin{algorithm}[t]
\caption{RotGuard}
\label{alg:rotguard}
\small
\begin{algorithmic}[1]
\Require Reference image $x$, prompt $c$,
  budget $\varepsilon$, step size $\alpha$, iterations $K$,
  EOT samples $M$, weights $w_{\mathrm{CLIP}},w_{\mathrm{Face}},w_Q,w_R$
\Ensure Protected image $x_p$
\State \textbf{Frozen:} SDXL U-Net $\epsilon_\theta$, IP-Adapter Plus Face
  token encoder $T$, CLIP image encoder $E_{\mathrm{CLIP}}$,
  FaceNet $E_{\mathrm{Face}}$
\State $\Pi(\delta)\!\gets\!\operatorname{clip}\!\big(x+\operatorname{clip}(\delta,-\varepsilon,\varepsilon),0,1\big)-x$
\State $G(V)\!\gets\!\hat V\hat V^{\top}$,
\State $y_c\gets T(x)$; \; $e_{\mathrm{CLIP}}\gets E_{\mathrm{CLIP}}(x)$; \;
  $e_{\mathrm{Face}}\gets E_{\mathrm{Face}}(x)$
\State $\delta\sim\operatorname{U}(-\varepsilon,+\varepsilon)$, \; $\delta\gets\Pi(\delta)$
\For{$k=1$ \textbf{to} $K$}
  \State $x_p\gets x+\delta$, \; $y_p\gets T(x_p)$
  \State $\mathcal L_{\mathrm{IFSR}}\gets
    w_{\mathrm{CLIP}}\big[1-\cos\big(E_{\mathrm{CLIP}}(x_p),e_{\mathrm{CLIP}}\big)\big]
    +w_{\mathrm{Face}}\big[1-\cos\big(E_{\mathrm{Face}}(x_p),e_{\mathrm{Face}}\big)\big]$
  \State $g\gets\nabla_\delta\,\mathcal L_{\mathrm{IFSR}}$
  \For{$m=1$ \textbf{to} $M$}
    \State Sample $t$,
      $z_0,\xi\sim\mathcal N(0,I)$
    \State $z_t\gets\operatorname{AddNoise}(z_0,\xi,t)$
    \State $F(y)\gets\epsilon_\theta(z_t,t,c,y)$
    \State $b\gets F(\mathbf 0)$
    \State $r_c\gets F(y_c)-b$;\;\;    record $Q_c^l,l=0.....L$
    \State $r_p\gets F(y_p)-b$;\;\;   record $Q_p^l,l=0.....L$
    \State $\mathcal L_{\mathrm{QSD}}\gets\frac1L\sum_{l}
      \|G(Q_p^l)-G(Q_c^l)\|_F^2$
    \State $\mathcal L_{\mathrm{CDS}}\gets\operatorname{MSE}\big(G(r_p),G(r_c)\big)$
    \State $g\gets g+\frac1M\nabla_\delta\big(w_Q\mathcal L_{\mathrm{QSD}}
      +w_R\mathcal L_{\mathrm{CDS}}\big)$
  \EndFor
  \State $\delta\gets\Pi\big(\delta+\alpha\operatorname{sign}(g)\big)$
\EndFor
\State \Return $x_p=x+\delta$
\end{algorithmic}
\end{algorithm}

\end{document}